\documentclass[runningheads]{llncs}

\usepackage[T1]{fontenc}
\usepackage{graphicx,verbatim}
\usepackage{color}

\RequirePackage{xspace}
\RequirePackage[dvipsnames]{xcolor}
\RequirePackage{graphicx}
\RequirePackage{amsmath}
\RequirePackage{amssymb}
\RequirePackage{booktabs}
\usepackage{xurl}
\usepackage{cite}

\usepackage{multicol}
\usepackage{multirow}
\usepackage{tabularray}
\usepackage{threeparttable}
\usepackage[table]{xcolor}
\definecolor{tblgray}{gray}{0.92}
\definecolor{ours}{HTML}{F3F7FC}
\definecolor{darkgreen}{rgb}{0.0,0.5,0.0}
\usepackage{cleveref}
\usepackage{graphicx,verbatim}
\usepackage{subcaption}

\definecolor{c2}{HTML}{ECF8F1}
\definecolor{c1}{HTML}{DEF3E6}
\newcommand{\cellfirst}[1]{\cellcolor{c1}\textbf{#1}}
\newcommand{\cellsecond}[1]{\cellcolor{c2}{#1}}

\newcommand{\pts}{\textcolor{darkgreen}{$\bullet$\,}}

\newcommand{\boldparagraph}[1]{\noindent\textbf{#1}}

\begin{document}
\title{Teeth2Point: A Two-Stage Dental CBCT ROI-to-Point Segmentation Framework}
\titlerunning{Teeth2Point}
%
\author{Qi Ma\inst{1,2}\orcidID{0009-0005-4028-6917} \and
Shipra Jain\inst{2}\orcidID{0009-0008-1833-3416} \and
Niko Benjamin Huber\inst{2}\orcidID{0009-0007-2089-0150}
\and
Ender Konukoglu\inst{1}\orcidID{0000-0002-2542-3611}} 

\authorrunning{Q. Ma et al.}
%
\institute{Computer Vison Lab, ETH Zurich, Switzerland 
\email{\{fname.lname\}@vision.ee.ethz.ch} \and
Align Technology
\email{\{sjain, nhuber\}@aligntech.com}
}


  
\maketitle              

\begin{abstract}
Modern deep learning architectures have demonstrated strong performance in dental CBCT segmentation. One remaining crucial challenge is accurate tooth labeling in cases with missing or malpositioned teeth, which are highly relevant for dental practice. Transformer-based architectures should in theory be able to resolve such ambiguities using global anatomical context. However, due to the high resolution of CBCT volumes and the wide spatial distribution of teeth within volumes, dense patch-based volumetric processing faces an inherent trade-off. Computational costs limit the number of patches that can be used in self-attention and thus, one can either increase the extent of the context captured in self-attention or capture fine-grained structural details by using small patches, but not both. 
In this work, we present Teeth2Point, an efficient point-based transformer framework for dental CBCT semantic segmentation that can avoid this trade-off. Teeth2Point first localizes volumetric regions of interest (ROIs) surrounding teeth using a convolutional model, then converts ROIs into point tokens using adaptive sampling. A transformer model predicts accurate segmentations using the point tokens, which allow capturing global context while retaining high resolution. 
The transformer is first pretrained using self-supervised learning (SSL), in the style of DINO but using domain-specific augmentation strategies, followed by supervised finetuning. The SSL pretraining, which includes random token masking, provides robustness to complex anatomical variations. 
Compared with the strongest two-stage baseline, Teeth2Point improves abnormal-case performance by 1.44 DSC points on average across four datasets; relative to the first-stage nnU-Net, the gain is 1.9 points.


\keywords{ 3D Medical Image Analysis  \and CBCT Semantic Segmentation \and Point Cloud Segmentation.}
\end{abstract}

\section{Introduction}
\label{sec:intro}

Volumetric Cone Beam Computed Tomography (CBCT) is widely used in dental practice and segmentation of such images is an important step for quantitative analysis and treatment planning. Advanced deep learning models have recently shown promising performance in this segmentation task. 
Despite the advances, consistent instance-level tooth labeling in difficult but relevant cases, particularly when teeth are missing or malpositioned, remains challenging. In such structural variations, using global anatomical context based on a large field of view, including bilateral symmetry and inter-arch relationships across the dentition, can help resolve ambiguities and improve robustness. However, using large context in volumetric segmentation with today's algorithms is challenging. 
Enlarging the context through downsampling sacrifices fine surface details required for accurate boundary segmentation whereas increasing the patch size can improve performance~\cite{toothfairy_2024TMI,isensee2021nnu} but incurs substantial computational cost for dense 3D volumetric processing. To improve scalability, prior work has explored hierarchical windowed attention \cite{hatamizadeh2022swinunetrswintransformers} and more recently Mamba-based state-space models \cite{U-Mamba}. Yet, processing large volumes remains expensive. This challenge is particularly evident in dental CBCT, where teeth occupy only a small fraction of the volume (as low as $0.5\%$, see Tab. \ref{tab:data_statistics}) while being widely dispersed across the scan.



Two-stage methods \cite{ho2021point,2018cascadeVNet,isensee2021nnu,lin2023accurate} address this sparsity by first extracting ROIs and then performing fine-grained segmentation in the ROIs. However, these approaches retain a dense volumetric representation, where background voxels remain dominant when foreground structures are spatially sparse, as in dental CBCT. In contrast, point-based representations focus on foreground structures and offer a promising alternative for balancing resolution and global context without excessive computation. However, existing volume-to-point methods \cite{ho2021point,He_2020} primarily rely on local feature modeling, limiting their ability to capture long-range anatomical relationships critical for instance-level tooth segmentation. Moreover, they do not leverage self-supervised pretraining to improve robustness in challenging regions with limited annotations.


To overcome these limitations, we propose \textbf{Teeth2Point}, a two-stage ROI-to-point segmentation framework that adopts transformer-based architecture with DINOv2-style self-supervised pretraining. During training, we first learn a foreground extraction model (e.g., nnU-Net~\cite{isensee2021nnu}) to compute the tooth ROI, which is expanded via dilation and converted into a point-based representation. The unlabeled data are used for self-supervised pretraining, followed by supervised fine-tuning separately on each benchmark dataset. In addition to CBCT-specific volumetric augmentations, we adopt adaptive sampling before tokenization, which allocates more points to high-detail surface regions. We progressively finetune the pretrained transformer, first on ground-truth foreground points and subsequently on predicted ROIs. During inference, the extracted foreground points undergo the same sampling procedure and are forwarded through the point-based network in a single pass to produce point-level predictions, which are subsequently interpolated and mapped back to the volumetric space.



We evaluate Teeth2Point on four dental CBCT datasets, achieving a 1.44 DSC-point gain in missing and malpositioned cases compared with the strongest two-stage baseline. Inference averages 14 seconds per volume, with the second stage running in under one second. Additionally, representations learned through self-supervised training improve label efficiency, achieving a Dice score of 86.7 using only 10 labeled samples.

\section{Related work}
\label{sec:related_work}

\begin{table}[t]
    \caption{\textbf{Dataset Characteristics.} The Region of Interest (ROI) refers to the cropped volume encompassing all foreground structures. Foreground Ratio (FR) denotes the proportion of foreground voxels in the entire volume, while FR-ROI refers to the foreground ratio within the region of interest. Abnormal Ratio (AR) measures the proportion of cases with missing or malpositioned teeth.} 
    
    \label{tab:data_statistics}
    \centering
    \fontsize{8}{9}\selectfont   
    \begin{threeparttable}
    \begin{tabular}{l|cccccc}
    \toprule[0.95pt]
    Dataset & Sample & Mean Shape & ROI Shape & FR & FR-ROI & AR  \\
    \midrule[0.6pt]
    STS2024\cite{cui2022ctooth+,wang2024sts} & 27 & (584,584,445)&(253,202,158) & 0.5$\%$ & 9.3\% & 82.5\% \\
    ToothFairy2\cite{toothfairy2_challenge_2026,toothfairy_2025CVPR,toothfairy_2024IEEEACCESS} & 480 & (386,348,182)& (214,169,113) & 1.4\% & 8.3\% & 80.8\% \\
    ToothFairy3\cite{toothfairy2_challenge_2026,toothfairy_2025CVPR,toothfairy_2024IEEEACCESS} &  532 & (386,349,181)&(215,170,112) & 1.3\% & 7.7\% & 23.1\% \\
    Internal Dataset & 1170 & (472,369,347) & (286,232,203) & 0.97\% & 4.3\% & 9.0\% \\
    \bottomrule[0.95pt]
    \end{tabular}
    \end{threeparttable}
\end{table}

\noindent \textbf{Volumetric Medical Image Segmentation.} CNN-based models such as nnU-Net \cite{isensee2021nnu} established strong performance in volumetric CT/MRI segmentation. Transformer-based methods (e.g., Swin UNETR, VideoTeeth) \cite{hatamizadeh2022swinunetrswintransformers,videoforteeth} improved global context modeling in 3D volumes, while hybrid architectures like MedNeXt \cite{roy2024mednexttransformerdrivenscalingconvnets} achieved strong benchmark performance. More recently, state-space models such as U-Mamba \cite{U-Mamba} used long-range context with linear complexity, making them suitable for dense 3D segmentation. Although these approaches show promising results for high-resolution dental CBCT segmentation \cite{cui2022ctooth+,toothfairy_2025CVPR,toothseg, umamba2}, performance often degrades in abnormal cases, where accurate instance-level labeling requires stronger global anatomical awareness.

\noindent \textbf{Two-Stage Framework and Volume-to-Point method.} Two-stage segmentation frameworks typically follow a coarse-to-fine strategy and are widely used in tumor \cite{2018cascadeVNet} and dental CBCT segmentation \cite{lin2023accurate}. 
Different than other organs, teeth are sparse and widely distributed and even in an ROI(defined as a cropped volumetric bounding box covering them), they constitute less than 10\% of the ROI volume (Tab.~\ref{tab:data_statistics}).
To exploit this property, we adopt a point-based representation in a second stage as in \cite{ho2021point,He_2020}. However, these approaches employ point-MLP-based networks \cite{qi2017pointnetdeephierarchicalfeature,hu2020randlanetefficientsemanticsegmentation}, which are mainly designed for local aggregation and provide limited modeling of global anatomical structure.



\begin{figure*}

    \centering   \includegraphics[width=\linewidth]{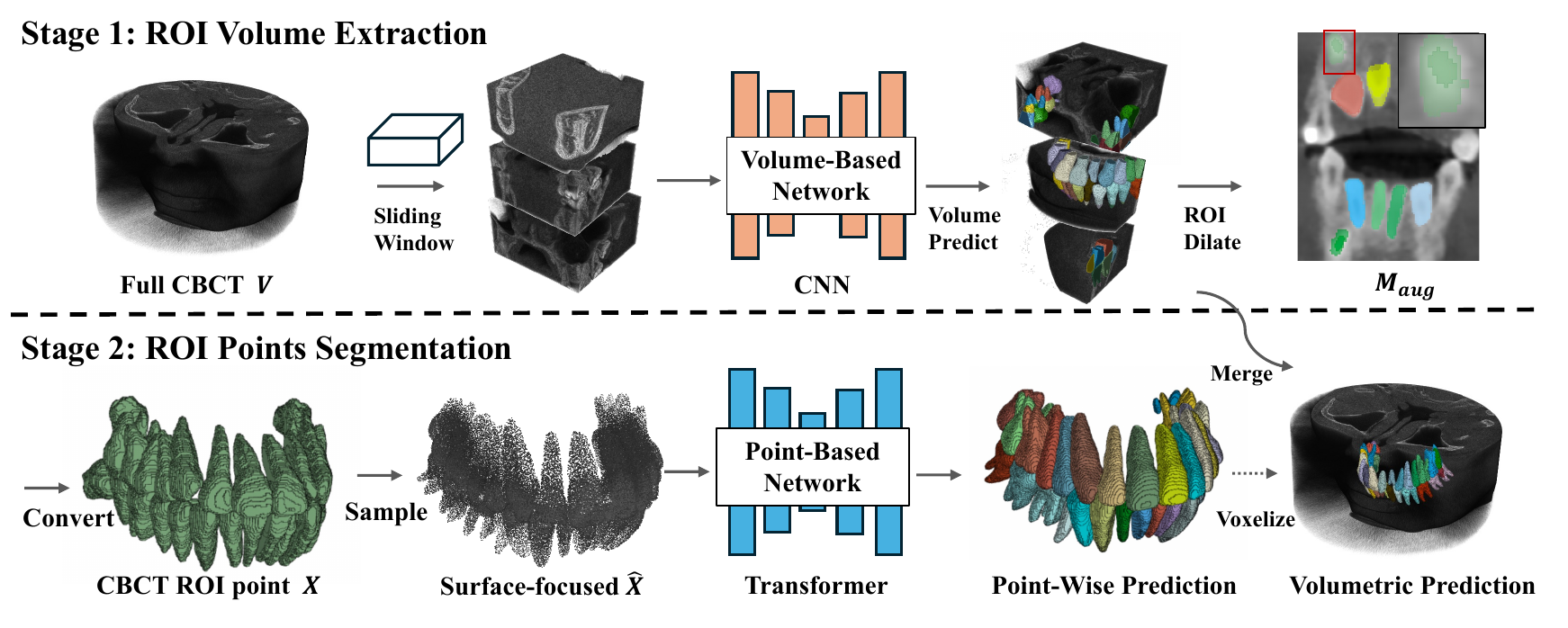}
    \caption{\textbf{Teeth2Point Overview}. The first stage uses a convolutional model for ROI extraction, followed by adaptive sampling to obtain a \textbf{sparse, surface-focused representation}. Point-wise predictions from the point-based transformer network are then mapped back to a volumetric form for evaluation.}
\label{fig:pipeline}
\end{figure*}


\noindent \textbf{Self-Supervised Pretraining for Volumetric Medical Modeling.}
SSL can reduce annotation requirements in volumetric segmentation. Beyond contrastive~\cite{chen2020simpleframeworkcontrastivelearning} and masked reconstruction~\cite{he2021maskedautoencodersscalablevision}, DINO-style self-distillation~\cite{oquab2024dinov2learningrobustvisual} learns strong semantic and global representations. However, its application to point-based representations in medical imaging remains underexplored. Motivated by recent advances shown for natural scenes \cite{wu2025sonata}, we adopt a DINOv2-style pretraining framework for dental CBCT point representations. We hypothesize that enforcing invariance to strong augmentations and masking encourages the model to rely on stable anatomical cues (e.g., bilateral symmetry and inter-arch spatial relationships), leading to more robust representations in challenging regions.

\begin{figure}[h]
    \centering
    \includegraphics[width=0.48\linewidth]{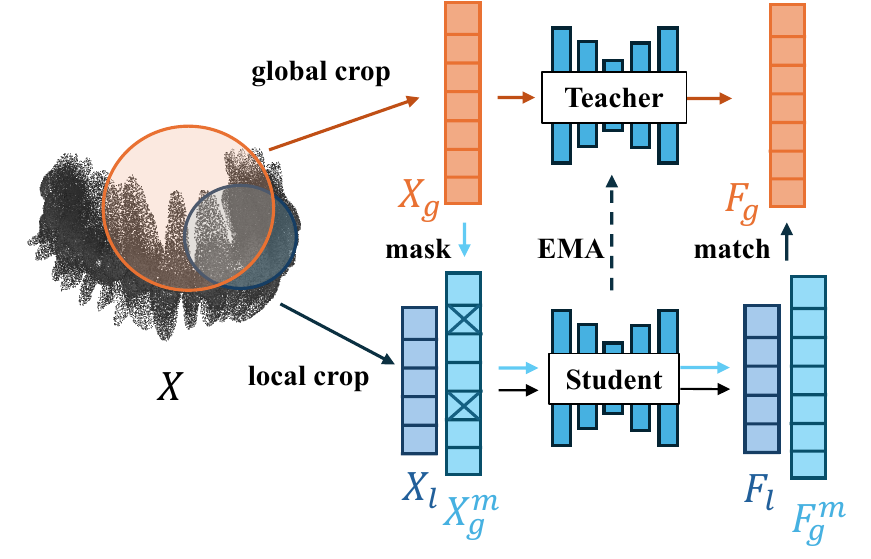}
    \hfill
    \includegraphics[width=0.48\linewidth]{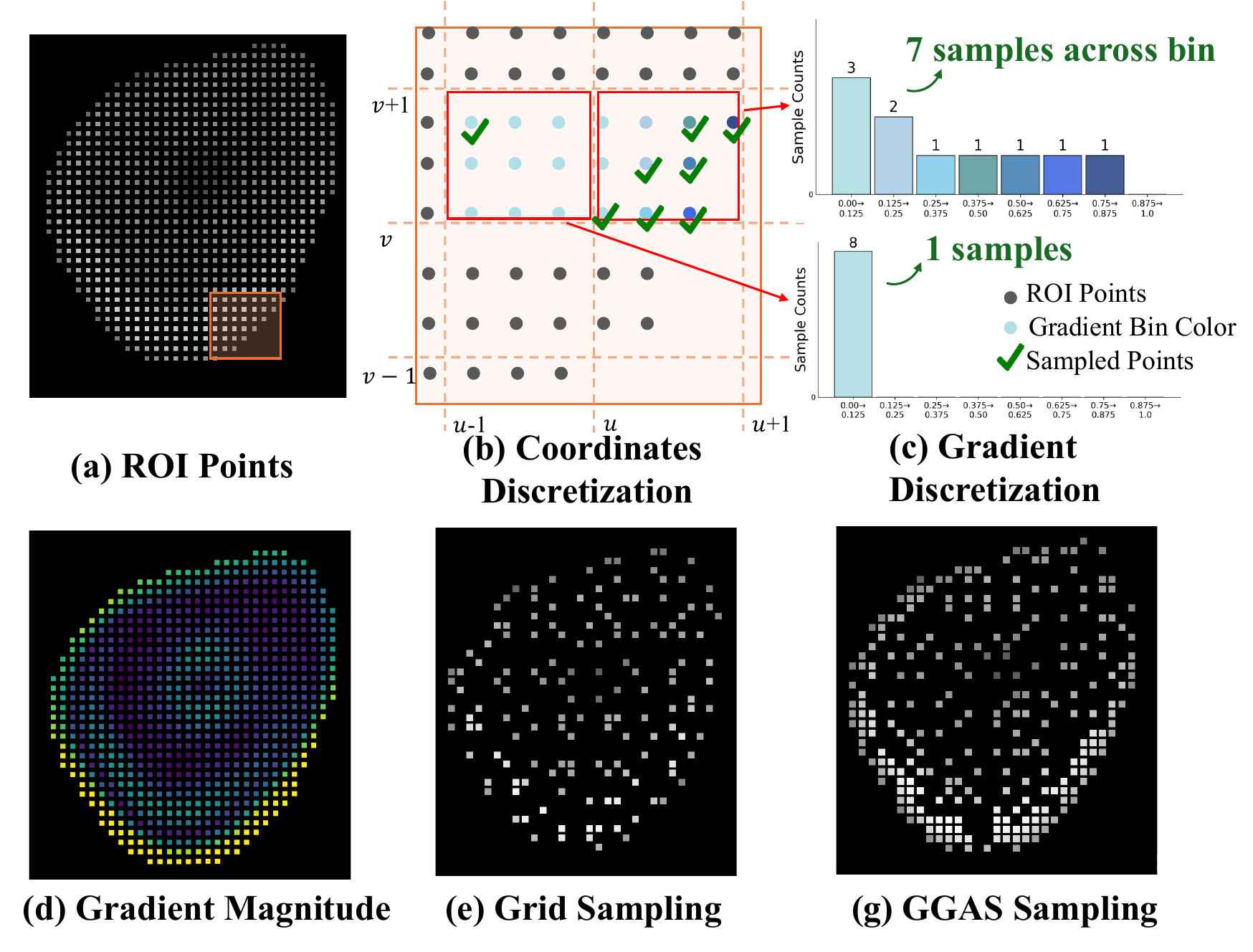}
    \caption{\textbf{Self-distillation pretraining and Gradient-Guided Adaptive Sampling illustration.} 
    DINOv2-style pretraining is applied to large-scale unlabeled ROI points (left), while GGAS allocates more samples to high-gradient boundary regions compared to uniform sampling (right).}
    \label{fig:ssl_and_gghs_illustrate}
\end{figure}

\begin{table*}[ht]
  \caption{\textbf{Quantitative Results}. We evaluate single- and two-stage methods on STS2024, ToothFairy2/3, and an internal dataset. nnU-Net results (marked with \pts) are used for second-stage ROI cropping. We report the Dice similarity coefficient (DSC) for all tooth classes and for molars separately, along with the average number of sliding windows (SW) and time for inference.}
  \label{tab:teeth_seg}
  \centering
  \resizebox{\linewidth}{!}{
  \begin{threeparttable}
  \begin{tabular}{lcc|cc|cc|cc|cc|cc}
  \toprule[0.95pt]
  \multirow{2}{*}{Method} & \multirow{2}{*}{SW} & Time  & \multicolumn{2}{c|}{\textit{STS2024}} & \multicolumn{2}{c|}{\textit{ToothFairy2}} & \multicolumn{2}{c|}{\textit{ToothFairy3}} & \multicolumn{2}{c|}{\textit{Internal}} & \multicolumn{2}{c|}{\textit{Average}} \\
  \cmidrule(lr){4-5} \cmidrule(lr){6-7} \cmidrule(lr){8-9}  \cmidrule(lr){10-11}  \cmidrule(lr){12-13}  
  &   &  (s) & DSC-R & DSC-A & DSC-R &  DSC-A & DSC-R &  DSC-A & DSC-R & DSC-A & DSC-R  & DSC-A \\
  \midrule[0.6pt]
  \rowcolor{tblgray}
  \multicolumn{13}{c}{\textit{Full-Volume CBCT}} \\
  \pts nnU-Net\cite{isensee2021nnu}  & 72 & 12.8 &  97.93  & 93.38   & 97.64 & 93.03 & 97.12 & 93.66 & 94.01 & 87.22 & 96.68 & 91.82   \\
    Toothseg\cite{toothseg}  & 72 & 28.3 & 96.76  & 93.05  & 98.46 & 93.22 & 98.08 & 91.96 & 93.93 & 87.17 & 96.81 & 91.35 \\

   MedNeXt\cite{roy2024mednexttransformerdrivenscalingconvnets} & 72  & 6.7 & 93.42 & 90.03  & 96.38 & 89.95 & 94.19 & 88.77 & 92.09 & 83.61 & 94.02 & 88.09  \\  
  VideoTeeth\cite{videoforteeth}  & 149 & 51.2 & 96.37 & 94.16  & 97.29 & 91.50 & 94.51 & 91.19 & 94.12 & 85.07 & 95.57 & 90.48    \\

  U-Mamba1\cite{U-Mamba}  & 75 & 33.9 & 96.87  & \cellsecond{94.47} & 98.21  & 92.17 & 95.54 & 94.20  & \cellfirst{94.22} & 86.85 & 96.21 & 91.92 \\

 U-Mamba2\cite{umamba2}  & 24 & 10.9 & \cellfirst{98.85} & 94.27  & \cellsecond{98.61} & 92.33 & 96.92 & 91.84 & 93.68 & 86.67 & 97.02 & 91.28 \\
    
  \rowcolor{tblgray}
  \multicolumn{13}{c}{\pts \textit{ROI-Cropped CBCT}} \\
   nnU-Net\cite{isensee2021nnu}  & 18 & 8.1 & 98.01 & 93.65 & 98.59 & \cellsecond{93.83} & \cellfirst{97.89} &  \cellsecond{94.27} & \cellsecond{94.13} & \cellsecond{87.27} & \cellsecond{97.16} & \cellsecond{92.26}  \\  
   Point-UNet\cite{ho2021point}   & 1   & 0.7 & 97.32  & 91.21 & 92.02 & 86.15 &  91.71 & 85.87 & 90.04 & 81.38 &  92.77 & 86.15 \\  
   Teeth2Point  & 1 & 0.9  &  \cellsecond{98.43} & \cellfirst{95.01} &  \cellfirst{98.93} & \cellfirst{95.17} & \cellsecond{97.84} & \cellfirst{95.86}  & 94.08 &  \cellfirst{88.74} &  \cellfirst{97.32} &  \cellfirst{93.70} \\  
  \bottomrule[0.95pt]
  \end{tabular}
  \end{threeparttable}
  }
\end{table*}

\section{Teeth2Point: Two-Stage Segmentation Framework}
\label{sec:method}

We illustrate the Teeth2Point method in Fig.~\ref{fig:pipeline} and provide details below.

\boldparagraph{CNN Based ROI Segmentation:} \label{subsec:first_stage} 
A CBCT volume is denoted as $V \in \mathbb{R}^{H \times W \times D \times 1}$. 
We first train a volumetric segmentation model $v_\phi$ to obtain an initial segmentation: $ \mathbf{O} = v_\phi(V) \in \mathbb{R}^{H \times W \times D \times (N+1)}, $
which contains logits for $N$ tooth classes and one background class. 
(Due to memory constraints, we use sliding-window inference and average logits in overlapping regions.) 
Voxel-wise predictions are obtained by $\mathbf{P}=\arg\max(\mathbf{O})$, and the foreground mask is defined as $\mathbf{M}=\mathbf{P}\neq \textrm{Background}$, i.e., all voxels that were assigned to a tooth by the CNN model. 
To obtain the final ROI, we further dilate the mask $\mathbf{M}_{\text{aug}}=\textit{Dilate}(\mathbf{M})$ using a 3D 6-connected (cross-shaped) kernel, expanding it by one voxel ($\approx 0.3 \text{mm}$) to avoid missing structures. The point-based representation $X$ is obtained by collecting all voxels inside the augmented mask.
Each point $\mathbf{x}_i \in X$ is associated with voxel coordinate $\mathbf{c}_i \in \mathbb{R}^3$, a normalized intensity $I_i$ (CBCT intensity clipped to $[-1000, 3512]$ and global z-score normalized), and intensity gradient vector $g_i$, forming the feature tuple $\mathbf{x}_i = (\mathbf{c}_i, I_i, g_i)$, and the point label $y_i$ from the corresponding voxel label for supervised fine-tuning.

\boldparagraph{Adaptive Sampling.} Processing all foreground points is computationally expensive, and discriminative cues for tooth segmentation are primarily located near structural boundaries. We therefore propose Gradient-Guided Adaptive Sampling (GGAS). Unlike uniform sampling, GGAS adapts sampling according to the local gradient distribution (Fig. \ref{fig:ssl_and_gghs_illustrate}, right). To this end, the gradient magnitudes of ROI points, $|g_i|$, are computed and normalized across all points via min-max normalization. We then create a coarser grid where each cell contains $\epsilon^3$ voxels. For each cell, a normalized gradient magnitude histogram is computed from ROI points that fall in the cell (Fig. \ref{fig:ssl_and_gghs_illustrate}c). Histograms are constructed using fixed-width bins (bin width $\alpha$). In each cell, one point is sampled from each non-empty histogram bin with a probability proportional to $|g_i|$. This produces $\hat{X}$ with more sampled points from cells with higher gradient magnitude variation and only one sample from cells with homogeneous gradient magnitude.

\boldparagraph{Architecture:} A linear embedding layer is used to convert point attributes into point tokens, followed by an encoder–decoder backbone \cite{wu2025sonata,wu2024ptv3}, which utilizes spatial pooling (max feature pooling within each voxel) and unpooling (nearest-neighbor interpolation) with skip connections for multi-level feature integration.

\boldparagraph{Self-Distillation Pretraining:} Self-distillation~\cite{oquab2024dinov2learningrobustvisual} learns augmentation invariant representations by training a student network $\theta_s$ to match an EMA teacher $\theta_t$~\cite{chen2020exploringsimplesiameserepresentation}. Given the point sets $X$, we construct two augmented views: a local view $X_l$ and a global view $X_g$, using domain-specific intensity and spatial transformations described further below. 
In addition, we apply random rotations and elastic transformations to the points, separately for local and global views. 
In case where adaptive sampling is not used, additional spherical neighborhood cropping is applied by randomly selecting a center point and sampling neighboring points according to a predefined ratio (local views: 0.1–0.4 and global views: 0.4–1.0).
As illustrated in \cref{fig:ssl_and_gghs_illustrate}, $\theta_t$ encodes $X_g$ into features $F_g$, while $\theta_s$ encodes $X_l$ into $F_l$. 
Instead of class tokens as in~\cite{oquab2024dinov2learningrobustvisual}, we utilize point-level distillation. Each point feature in $F_l$ is associated with its nearest neighbor in $F_g$; these paired features are converted into probabilities, and $\theta_s$ is optimized by minimizing the KL-divergence at point-level.
Following~\cite{oquab2024dinov2learningrobustvisual}, we stabilize training with Sinkhorn-Knopp centering and clustering-based assignment normalization~\cite{caron2021unsupervisedlearningvisualfeatures}, which help avoid representation collapse and encourage discriminative embeddings. 
Moreover, we construct a masked global view $X_g^m$ by randomly masking points with a training-scheduled ratio $r$~\cite{zhou2022ibotimagebertpretraining}. 
We insert a learnable mask token at the masked locations and $\theta_s$ embeds the masked points to $F_g^m$.
In addition to the point-level distillation, $\theta_s$ is also trained to minimize the difference between $F_g$ and $F_g^m$, using the same objective as the distillation. 
This masked prediction task encourages the model to capture richer contextual geometry and learn robust point representations.
%
\boldparagraph{Volumetric Medical Domain Augmentation:} We adapt commonly used volumetric CBCT augmentations to point attributes $(\mathbf{c}_i, I_i, g_i)$ during training. Intensity transformations (e.g., gamma correction and brightness scaling) perturb per-point intensities. We further introduce two dental-specific strategies: \textit{Symmetry Augmentation}, which flips points with label remapping to exploit bilateral tooth structure, and \textit{Artifact Augmentation}, which simulates metal-induced streak artifacts common in clinical CBCT by randomly selecting a tooth class and amplifying the intensities of its associated points.

\begin{figure}[!b]
    \centering   
    \includegraphics[width=\linewidth]{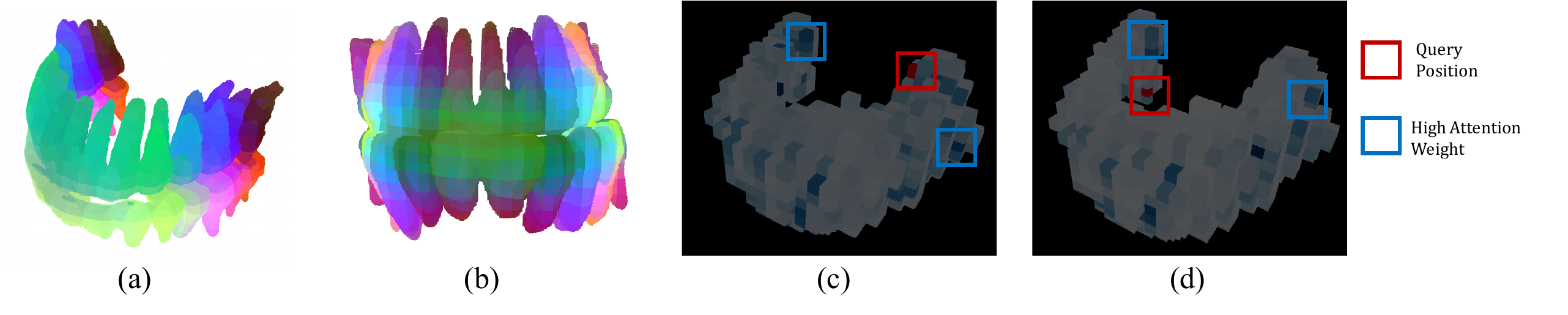}
   \caption{\textbf{Visualization of pretrained representations and attention weights.} We show PCA projections of feature dimensions 1--3 (a) and 4--6 (b), along with query--key attention scores in (c) and (d).}
\label{fig:attention_vis}
\end{figure}


\boldparagraph{Progressive finetuning:} A two-step strategy is employed for supervised fine-tuning, where both stages optimize a combined Dice and cross-entropy loss with equal weighting. In the first (GT-ROI) stage, labels contain only foreground points. In the second adaptation stage, the input is derived from predicted ROIs that may be inaccurate; consequently, the corresponding point labels, inherited from voxel annotations, include background and incomplete structures.




\boldparagraph{Voxelization:} The model produces point-wise predictions on the sampled points. Predictions are propagated to unsampled points using K-nearest neighbors interpolation. The point-wise predictions are then voxelized into a volumetric representation and combined with background labels from the first-stage results, i.e., where $M_\textit{aug}(\mathbf{p}) = 0$.



\section{Experiments}
\label{sec:experiments}
\noindent\textbf{Implementation Details.} All volumes have an isotropic spacing of $0.3~\text{mm}$. We set $\epsilon=2$ for coordinate discretization, $\alpha=0.125$ for gradient histogram binning, and use $K=4$ for prediction voxelization. For pretraining, we use 3,224 internal unlabeled CBCT scans with a batch size of 32 and a cosine schedule that increases the mask ratio from $0.3$ to $0.7$, trained for 200 epochs on 8 NVIDIA A6000 GPUs. Fine-tuning is performed with a batch size of 8 for 400 epochs GT-ROI training, followed by 100 adaptation epochs on a single A6000 GPU, conducted separately for each benchmark dataset. All supervised comparisons use the same 70/30 train-test split within each dataset and small-component removal post-processing \cite{isensee2024scalingnnunetcbctsegmentation}. Because this short workshop paper reports one split, the gains should be interpreted as descriptive rather than inferential.

\begin{figure*}[!t]
    \centering
    \includegraphics[width=\linewidth]{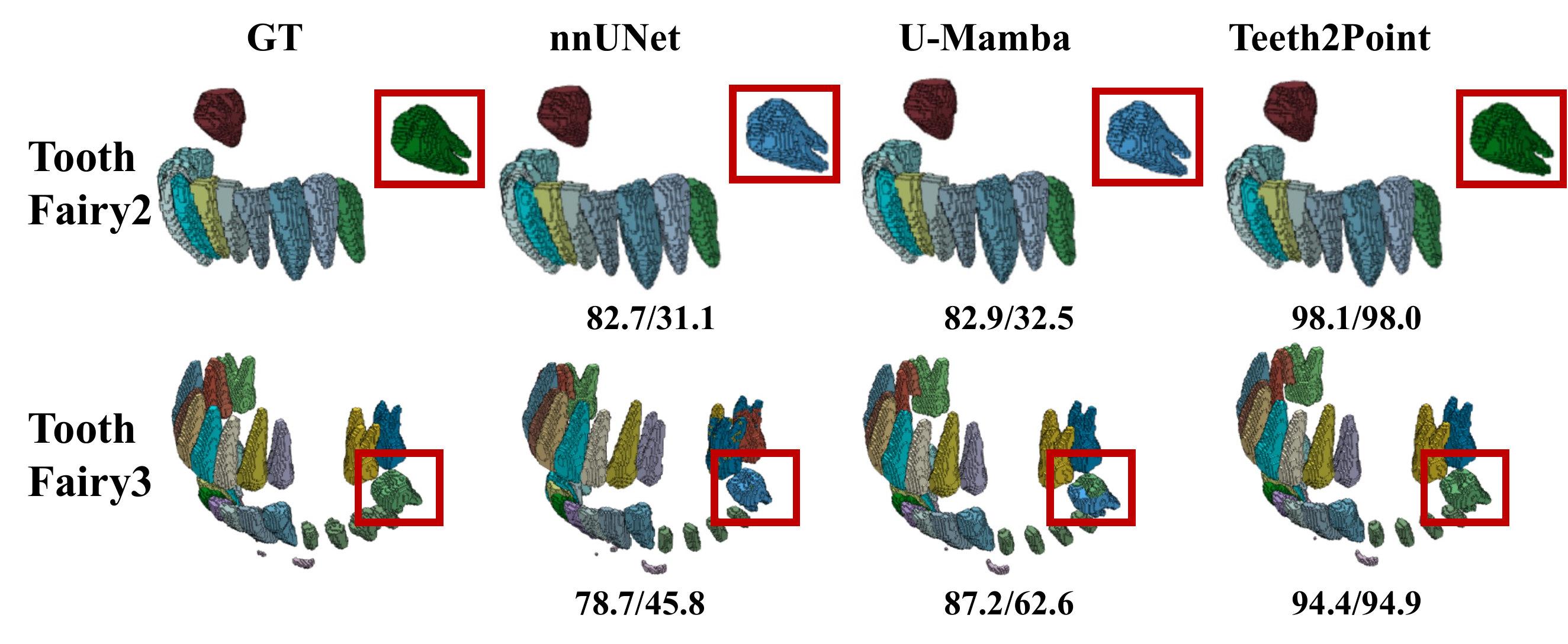}
    \caption{\textbf{Qualitative Results.} Teeth2Point shows strong performance among two-stage methods (numbers denote Average DSC / Molar DSC).}
    \label{fig:qualitative_results}
\end{figure*}

\boldparagraph{Main Results:}
We compare Teeth2Point against full-volume and ROI-cropped nnU-Net baselines~\cite{isensee2021nnu} and other single- and two-stage baselines (Tab.~\ref{tab:teeth_seg}). These include ToothSeg and MedNeXt~\cite{roy2024mednexttransformerdrivenscalingconvnets} (CNN-based), VideoTeeth~\cite{videoforteeth} (transformer-based), and U-Mamba 1\&2~\cite{U-Mamba,umamba2} (state-space model). We report DSC-R and DSC-A separately for regular and abnormal cases, where all baselines show a clear performance drop, with scores \textbf{5.2\%} below the overall average. The ROI-cropped nnU-Net improves over the full-volume nnU-Net, potentially because it focuses more on the foreground region, whereas Point-UNet performs comparatively worse, as RandLA-Net \cite{hu2020randlanetefficientsemanticsegmentation} primarily emphasizes local feature aggregation. Teeth2Point achieves the best average DSC-A and improves upon the strongest ROI-cropped baseline by \textbf{1.44} DSC points. It also improves upon the first-stage nnU-Net by \textbf{0.6} points in DSC-R and \textbf{1.9} points in DSC-A, reducing the regular-abnormal performance gap by approximately 26\%. As shown in Fig.~\ref{fig:attention_vis}, the pretrained representations capture pronounced left-right symmetry (a) and upper-lower structural relationships (b). The final encoder attention maps further show that molar-region tokens attend to anatomically corresponding counterparts across horizontal and vertical directions (c-d). These results support the narrower interpretation that Teeth2Point can access global anatomical context, while causal attribution of the gain to remote context requires additional controlled experiments. The point stage is efficient, requiring one pass and 0.9 seconds, while the full pipeline requires approximately 13.7 seconds including first-stage ROI extraction. We were unable to submit to the Grand Challenge leaderboard as the provided NVIDIA T4 GPUs lack FlashAttention-2 support required to meet the memory demands of large-scale token attention.



\begin{figure}[!t]
    \centering   \includegraphics[width=
    \linewidth
    ]{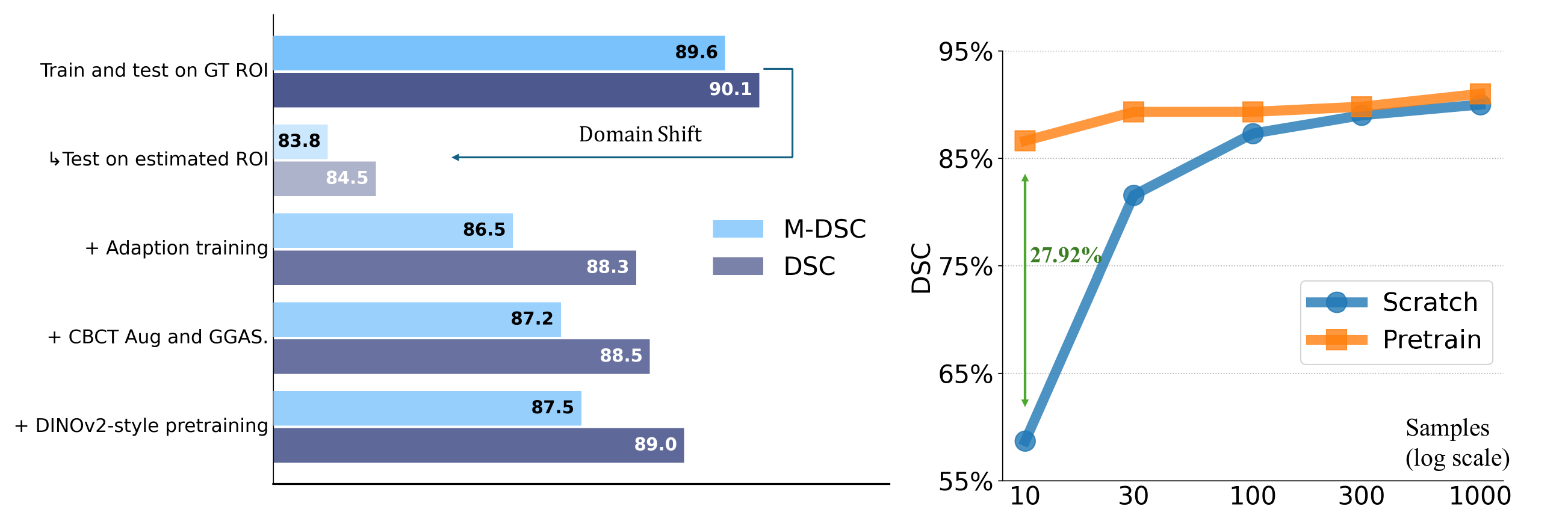}
    \caption{\textbf{Experiments of Ablation (left) and Label Efficiency (right).} }
\label{fig:ablation_design}
\end{figure}


\boldparagraph{Ablation.} Teeth2Point was initially trained on point clouds extracted from GT ROIs. When evaluated on points from predicted ROIs, the DSC decreased by over $5.6\%$ on the internal dataset (Fig.~\ref{fig:ablation_design} left). Fine-tuning on imperfect predictions with adapted CBCT augmentation and gradient-guided sampling largely recovered the performance loss and notably improved results in the molar region. In addition, the pretrained model further contributed to the overall performance.

\boldparagraph{Label efficiency.} We report the effectiveness of the pretrained model in Fig.~\ref{fig:ablation_design} (right). The pretrained model consistently outperforms training from scratch under the same subset protocol. Notably, with only \textbf{10} labeled samples for fine-tuning, it achieves an average DSC of \textbf{86.7\%}, which is comparable to training from scratch with 100 samples. This suggests reduced annotation needs, although repeated subset draws would be required to quantify uncertainty.





\section{Conclusion}
\label{sec:conclusion}
We present Teeth2Point, a two-stage framework that converts dense CBCT volumes into point-based representations and processes them with a self-distillation pretrained transformer. Several design components improve robustness and global context modeling. Experiments show improved performance, particularly in abnormal cases. A current limitation is that the framework is not fully end-to-end, as it relies on a separate ROI extraction model in the first stage; if a tooth is entirely missed during ROI extraction, the second stage cannot recover it.

\bibliographystyle{splncs04}
\bibliography{camera_ready}
%




\end{document}